\documentclass[10pt]{article}

\usepackage[utf8]{inputenc}
\usepackage[T1]{fontenc}
\usepackage{microtype}
\usepackage[a4paper, margin=2.4cm]{geometry}
\usepackage{lmodern}
\usepackage{graphicx}
\usepackage{booktabs}
\usepackage{tabularx}
\usepackage{xcolor}
\usepackage{amsmath, amssymb}
\usepackage{hyperref}
\usepackage{cleveref}  
\usepackage{tikz}
\usetikzlibrary{positioning, arrows.meta, shapes.geometric}
\usepackage{listings}
\usepackage{titlesec}
\usepackage{enumitem}
\usepackage{placeins}  

\hypersetup{
  colorlinks=true,
  linkcolor=black,
  citecolor=blue!60!black,
  urlcolor=blue!60!black
}

\titleformat{\section}{\Large\bfseries}{\thesection}{0.5em}{}
\titleformat{\subsection}{\large\bfseries}{\thesubsection}{0.5em}{}
\setlist[itemize]{leftmargin=1.5em, topsep=2pt, itemsep=1pt}

\title{\bfseries Moxia: A Trust-First Neuro-Symbolic Execution\\
       Architecture for Self-Explaining Mathematical Reasoning}

\author{%
  Alessio Bruno\thanks{Independent researcher.
  Correspondence: \texttt{alessiobruno1992@gmail.com}}
}
\date{\today}

\begin{document}
\maketitle

\begin{abstract}
\noindent
We present \textsc{Moxia}, a trust-first neuro-symbolic execution
architecture for self-explaining mathematical reasoning over
natural-language input. In
\textsc{Moxia}, the language model functions strictly as a
\emph{canonicalizer}: it rewrites informal problem text into a narrow
schema consumed by a deterministic Computer-Algebra-System (CAS)
pipeline, which derives and verifies the answer or abstains as a
first-class output. Routing follows a $1{:}1{:}1$ alignment between
problem-shape regex, schema-specific prompt, and closed-form CAS
handler, with $5{,}741$ such routes shipped, $75.9\%$ of which reach
an answer without invoking the language model at all, and zero
$\textsc{lost\_correct}$ regressions as a standing release gate.

Because the answer is \emph{derived} rather than generated, the
derivation itself is available: every handler emits a step trace from
the computation it actually performed, and a synthesizer renders it as
prose; where the derivation fits a fragment of Lean~4 with Mathlib, the
same declared data become a theorem the kernel checks. Every task carries
an explicit decision on both layers, and neither invents a step.

On the $5{,}000$-record \textsc{MATH}\ test split across all $7$
categories, shapes the task registry was designed against,
\textsc{Moxia}\ answers $95.6\%$ correctly ($4{,}780/5{,}000$) with
\emph{zero} confident-wrong answers and abstains, saying where, on the
other $220$; the \textsc{MATH}-500 subset of that split reads $95.2\%$
($476/500$), also with zero wrong. We release the run as a public
file, one record per problem: the problem, the expected answer, the
model's transcription, Moxia's answer or its abstention, the
explanation, and the Lean theorem, marked record by record as
compiled. $3{,}142$ of the $4{,}780$ correct answers ($65.7\%$) carry a
theorem, and all $3{,}142$ compile. Building the theorems actually
served, rather than only a fixture corpus, is a gate of its own: its
first pass found seven that did not compile, two of them false
statements beside correct answers, all repaired before release.

The contribution we emphasize is not a final accuracy figure but the
\emph{forward dynamic} the architecture establishes: every logged
abstain in production is a candidate correct after one ship cycle,
since new tasks compose without regressing the registry. The
operational discipline behind this property (math-template
bucketing, $\textsc{lost\_correct}$ scan as regression oracle,
parseable-first onboarding, and abstain as first-class output)
constitutes a transferable framework for trustworthy neuro-symbolic
systems beyond mathematics.
\end{abstract}

\section{Introduction}
\label{sec:intro}

\paragraph{The unverifiable-\textsc{LLM}\ math problem.}
Frontier large language models achieve impressive accuracy on
mathematical reasoning benchmarks, but they expose no
verification pathway: at the \textsc{API}\ level, a
confident-wrong answer is indistinguishable from a confident-right
one. The user has no structural recourse to know whether a given
output is reliable. This is not a defect of any particular model;
it is a structural property of the
\textsc{prompt-in-text-out} interface itself.

Two existing alternatives partially address verification but at
the cost of severely restricting input scope. Lean-based provers
with \textsc{LLM}\ copilots~\cite{song2024leancopilot,
azerbayev2023llemma} verify each tactic against the Lean kernel
but require the problem to be pre-formalized in Lean syntax;
the formalization is itself the bottleneck for natural-language
queries. Closed expert systems such as Wolfram
Alpha~\cite{wolfram-alpha} answer \textsc{NL}\ input with rich
symbolic backends, but their derivation traces are not inspectable
and the system is not \textsc{LLM}-augmented at the input boundary.

\paragraph{Trust-first vs accuracy-first.}
We define \emph{trust} as $1 - \text{wrong}/\text{attempted}$,
where \emph{wrong} excludes records on which the system explicitly
returned \textsc{unknown}. Trust is distinct from \emph{accuracy}
($\text{correct}/\text{total}$): a system that refuses unsafe
questions can have very high trust even at modest accuracy. Our
position is that \emph{confident-wrong is the worst failure mode}
in mathematical reasoning, and an architecture should be designed
to make confident-wrong structurally rare, not to be punished
\emph{post hoc} by benchmarks.

\paragraph{Bottom-up architectural commitment.}
\textsc{Moxia}\ commits to four design choices that follow from
the trust-first stance: (a)~the language model serves as a
\emph{canonicalizer} that rewrites \textsc{NL}\ input into a narrow
task-specific schema, never as a solver; (b)~a deterministic
\textsc{CAS}\ pipeline derives every emitted answer; (c)~routing
between the \textsc{LLM}\ and the \textsc{CAS}\ pipeline is
template-aligned: each routed task is one
(trigger, prompt, handler) triple
co-designed against the same math template;
and (d)~\textsc{abstain} is a first-class structural output emitted
by any of three independent channels (no template match, \textsc{LLM}\
\textsc{unknown}, handler cannot derive). The combination yields
a runtime trust guarantee that is not available to either
monolithic \textsc{LLM}\ systems or pre-formalized provers.

\paragraph{Contributions.}
\begin{itemize}
  \item \textbf{Architecture} (\Cref{sec:arch}): a $1{:}1{:}1$
        (trigger, prompt, handler)
        routing architecture, an operator-pipeline chain framework
        for multi-step shapes, and a \texttt{rule\_only}
        \textsc{LLM}-bypass for closed-form shapes. We ship
        $5{,}741$ task triples at the time of writing, $4{,}356$
        of them ($75.9\%$) on the \texttt{rule\_only}
        path.\footnote{The task tree holds $5{,}743$ files. Two of
        them define a task that is not wired into the router, so the
        \emph{routable} count is $5{,}741$; counts stated per task
        file (the Lean export of \Cref{sec:empirical:lean}) therefore
        read $5{,}743$.}
  \item \textbf{Faithful derivations by construction}
        (\Cref{sec:arch:explain}): because the answer is derived by a
        \textsc{CAS}\ handler rather than generated, the explanation is
        rendered \emph{from that derivation}, not composed by a
        language model. Every routed task emits one, and it cannot
        narrate a step the handler did not take.
  \item \textbf{Machine-checked answers, record by record}
        (\Cref{sec:empirical:lean}): every task carries an explicit
        Lean~4 decision, a theorem or a stated abstention; the corpus
        holds $7{,}567$ theorems, and on the test split $3{,}142$ of
        the $4{,}780$ correct answers carry one, each compiled in the
        release's own build.
  \item \textbf{Empirical evaluation} (\Cref{sec:empirical}):
        $95.6\%$ correct ($4{,}780/5{,}000$) on the full $7$-category
        \textsc{MATH}\ test split with zero confident-wrong answers, and
        $95.2\%$ ($476/500$) on its \textsc{MATH}-500 subset, also with
        zero. Every domain sits above $92\%$. The \texttt{rule\_only}
        path achieves $100\%$ on the $20{,}000$-record lm-eval-harness
        arithmetic suite.
  \item \textbf{A public, per-record release}
        (\Cref{sec:availability}): the whole test run as one file,
        with the problem, the gold, the transcription, the answer or
        abstention, the explanation and the theorem, so that every
        claim above can be checked record by record, and every theorem
        rebuilt, without our code.
  \item \textbf{Operating principles} (\Cref{sec:discussion}):
        four principles transferable beyond mathematics (
        math-template bucketing, \textsc{LOST\_CORRECT} scan as
        regression oracle, predicate-not-recognized as mandatory
        abstain, and parseable-first onboarding with
        regime-dependent trust floor), each justified by direct
        empirical observation across hundreds of ship cycles.
\end{itemize}

\begin{figure}[p]
  \centering
  \includegraphics[width=0.62\linewidth,height=0.86\textheight,%
                   keepaspectratio]{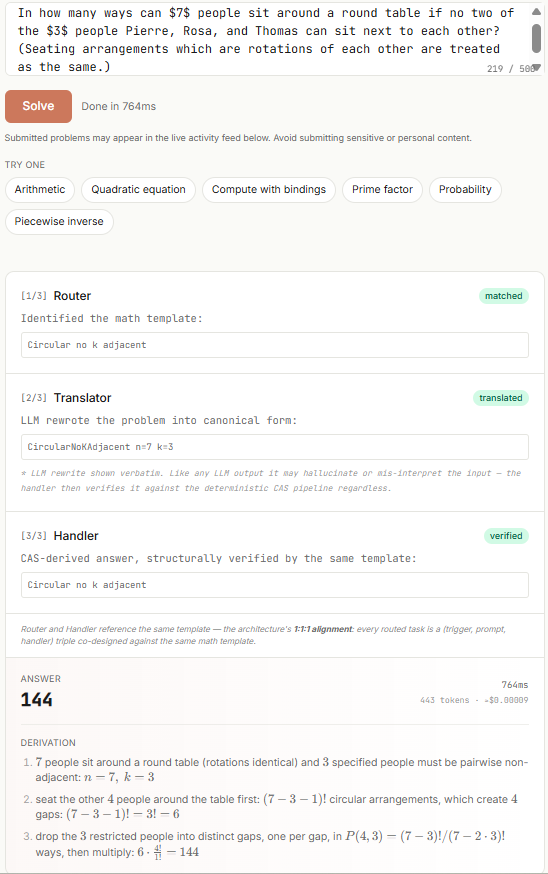}
  \caption{Single-query trace from the production demo (August 2026) on
    a circular seating problem: \emph{in how many ways can $7$ people sit around a
    round table if no two of Pierre, Rosa and Thomas are adjacent?}
    The three numbered pipeline stages (Router, Translator, Handler),
    followed by the answer and its derivation, materialize the
    $1{:}1{:}1$ alignment of \Cref{sec:arch:1to1to1}: Router and
    Handler name the same \emph{Circular no $k$ adjacent} template.
    The division of labour is legible throughout: the \textsc{LLM}'s
    entire contribution is the canonical
    \texttt{CircularNoKAdjacent n=7 k=3}, two integers transcribed
    from the prose, and no mathematics. Every mathematical step below is
    the handler's. The \textsc{Derivation} block is rendered from that
    handler's own trace (\Cref{sec:arch:explain}), which is why it can
    exhibit the actual argument (seat the $4$ unrestricted people in
    $(7-3-1)!$ circular arrangements, creating $4$ gaps, then place the
    $3$ restricted people into distinct gaps in $P(4,3)$ ways, giving
    $6\cdot 24=144$) rather than a plausible narration of it. The
    answer reports the per-query cost ($443$ tokens
    $\approx$\,$\$0.00009$ at Together.ai\ \$0.18/M;
    \Cref{sec:empirical:token-efficiency}). Any of the three abstain
    channels (router miss, \textsc{LLM}\ \textsc{unknown}, handler
    abstain) becomes visible at the stage that emitted it.}
  \label{fig:demo-trace}
\end{figure}

\begin{figure}[t]
  \centering
  \includegraphics[width=0.92\linewidth]{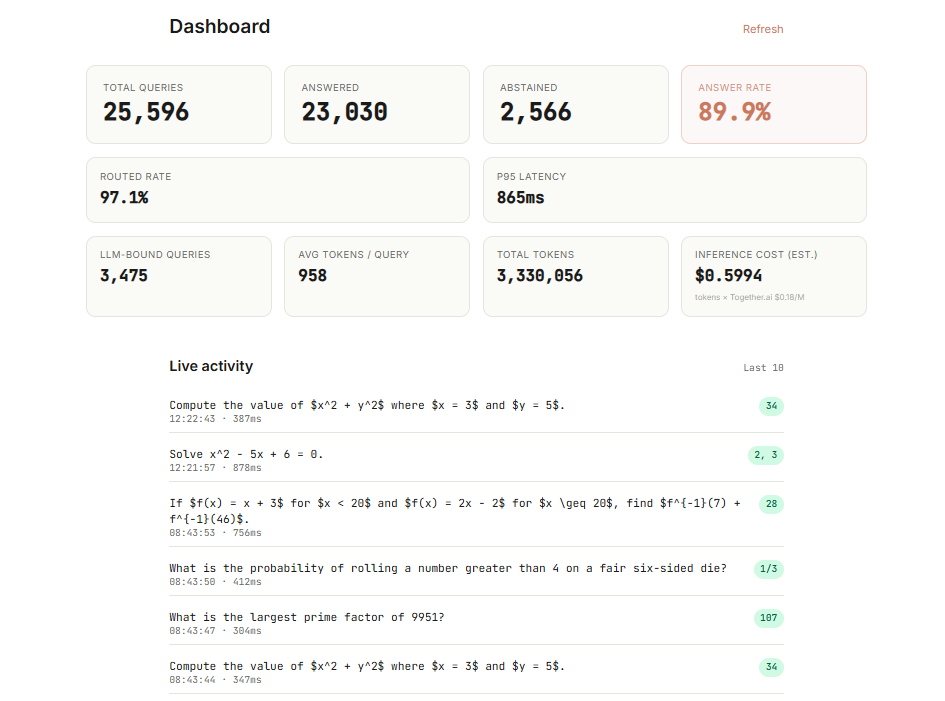}
  \caption{Cumulative dashboard from the production deployment.
    Top row: hero counters (Total queries, Answered, Abstained,
    Answer rate). Middle row: structural correctness signals
    (Routed rate $\sim 97\%$; p95 latency $\sim 865\,$ms dominated
    by \textsc{LLM}-bound traffic). Bottom row: efficiency aggregates
    confirming the per-query footprint of
    \Cref{sec:empirical:token-efficiency} (LLM-bound queries,
    average tokens / query, total tokens, inference cost
    estimated as tokens $\times$ Together.ai \$0.18/M). The live
    activity log (below) anchors the dashboard to real,
    free-form user traffic; each entry is one query, with
    per-record latency and routed task.}
  \label{fig:demo-dashboard}
\end{figure}

\paragraph{Reproducibility.}
A live, publicly accessible deployment of the architecture runs at
\url{https://huggingface.co/spaces/Squagghy/moxia}. The
single-query view (\Cref{fig:demo-trace}) illustrates the
$1{:}1{:}1$ alignment on every record and the structural visibility
of the abstain channels; the cumulative dashboard
(\Cref{fig:demo-dashboard}) exposes the production statistics
referenced in~\Cref{sec:empirical:production}. The test-split run
behind every figure of \Cref{sec:empirical} is released record by
record (\Cref{sec:availability}). Every number in this paper is
measured with Qwen~2.5 7B Instruct; the public deployment now serves a
different model of the same family, because its hosting provider no
longer offers the 7B, so answers given by the demo may differ from the
released run.

\FloatBarrier

\section{Architecture}
\label{sec:arch}

\begin{figure}[t]
\centering
\resizebox{\linewidth}{!}{%
\begin{tikzpicture}[
  node distance=1.4cm,
  every node/.style={font=\footnotesize},
  block/.style={
    rectangle, draw, rounded corners=2pt, minimum width=2.4cm,
    minimum height=0.9cm, align=center, fill=gray!10
  },
  llm/.style={
    rectangle, draw, rounded corners=2pt, minimum width=2.4cm,
    minimum height=0.9cm, align=center, fill=blue!8
  },
  arrow/.style={-{Latex[length=2mm]}, thick},
]
  \node[block] (input) {Problem text\\(natural language)};
  \node[block, right=of input] (router) {Router\\(regex, $O(1)$)};
  \node[llm, right=of router] (llm) {LLM rewriter\\(canonicalizer)};
  \node[block, right=of llm] (handler) {Handler\\(CAS, deterministic)};
  \node[block, right=of handler] (out) {Answer\\or \textsc{unknown}};

  \draw[arrow] (input) -- (router);
  \draw[arrow] (router) -- node[above]{\scriptsize task} (llm);
  \draw[arrow] (llm) -- node[above]{\scriptsize schema} (handler);
  \draw[arrow] (handler) -- (out);

  \draw[arrow, dashed]
    (router.south) to[bend right=22] node[below, midway]{\scriptsize rule\_only=True}
    (handler.south);
\end{tikzpicture}%
}
\caption{\textsc{Moxia}\ pipeline. The router selects exactly one
task per query (regex match on problem-shape, $O(1)$). The LLM
rewrites the input into a task-specific schema; the handler
deterministically derives and verifies the answer via SymPy. The
\texttt{rule\_only=True} path bypasses the LLM for math-template-pure
shapes (e.g.\ bare arithmetic; $88\%$ of lm-eval arithmetic records).}
\label{fig:pipeline}
\end{figure}
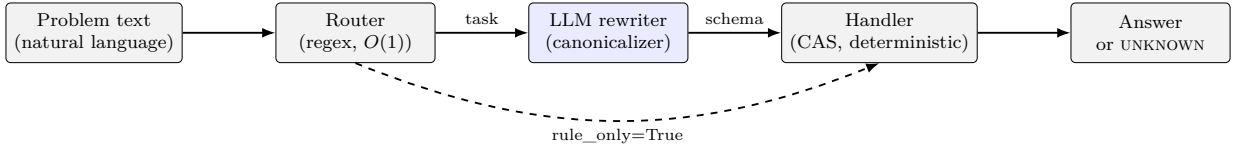

The architecture realizes the trust-first stance from
\Cref{sec:intro} as a deterministic execution path: a problem-shape
regex selects exactly one task, the language model rewrites the
input into that task's narrow schema, and a closed-form
\textsc{CAS}\ handler derives and verifies the answer or abstains
through a structured fail-reason. The four design choices we
enumerate below, $1{:}1{:}1$ task routing alignment
(\Cref{sec:arch:1to1to1}), abstain as a first-class output
(\Cref{sec:arch:abstain}), the composed-task chain framework for
multi-step shapes (\Cref{sec:arch:chain}), and the
\texttt{rule\_only} path that bypasses the \textsc{LLM}\ entirely
on math-template-pure shapes (\Cref{sec:arch:rule-only}), are
each motivated by a specific failure mode of the
\textsc{prompt-in-text-out} interface. \Cref{fig:pipeline}
summarizes the execution path of a single query.

\subsection{\texorpdfstring{$1{:}1{:}1$}{1:1:1} task routing alignment}
\label{sec:arch:1to1to1}

Routing in \textsc{Moxia}\ departs from the two prevailing patterns
in \textsc{LLM}-augmented mathematical reasoning: ``one model emits
the answer end-to-end'' (frontier monolithic systems) and ``the
\textsc{LLM}\ rewrites into a single structured form, then a generic
\textsc{CAS} handler consumes it'' (early hybrid systems). Instead,
every problem shape we cover is carved out as a \emph{triple} of
(a)~a regex trigger, (b)~a prompt whose few-shot examples teach a
schema specific to that shape, and (c)~a deterministic handler that
consumes only that schema. The three components are co-designed so
the trigger fires precisely on shapes whose canonical the prompt can
produce and whose answer the handler can verify. We call this the
\emph{$1{:}1{:}1$ alignment invariant}: one trigger, one prompt, one
handler.

The invariant is load-bearing in two senses. First, \emph{trust
attribution becomes local}: a record routed to task $T$ that produces
an answer is verifiable strictly through $T$'s code path; no
emergent behaviour spans tasks. Second, \emph{registry growth is
linearly additive}: adding task $T_{N+1}$ cannot regress tasks
$T_{1..N}$ because their code paths are disjoint by construction
(non-overlapping triggers, isolated handlers). Across the
$N=5{,}741$ task triples shipped at the time of writing,
$\textsc{lost\_correct}$ regressions cumulated to $0$ across the
release history, apart from a handful of records moved to abstention
on purpose because the answer that had been credited was a coincidence
(a truncated or ill-posed gold). A pre-commit migration scan against
archived bench \textsc{JSON} files acts as the regression oracle:
any trigger widening or new task that would steal a previously-correct
record and produce a worse outcome is caught before commit
(see~\Cref{sec:discussion}, Principle~\#17).

This stands in deliberate contrast with monolithic
\textsc{LLM}{}-as-solver systems, in which each new capability shares
representational budget with all existing ones, and adding capability
$N+1$ implicitly competes against $1..N$ for prompt space, attention,
and retrieval relevance.

\paragraph{What the model writes.}
On most routes the model emits one line, the canonical. On a declared
set of word-problem prompts it is allowed to work first: it names each
quantity the prose gives and what it refers to, and only then writes
the canonical on its last line, under the instruction not to compute
the final answer itself, so that the canonical still \emph{contains}
the arithmetic rather than its result. Only that last line reaches the
handler. The working is a transcription aid (the measured failures of
small models on word problems are binding failures, which number
belongs to which entity, not arithmetic ones), and it does not make the
model a solver: whatever it writes before the canonical, the answer is
the handler's.

\subsection{Abstain as first-class output}
\label{sec:arch:abstain}

A response with \texttt{answer=null} is structurally distinct in
\textsc{Moxia}\ from a response with \texttt{answer=value}. Three
independent channels feed the same null:

\begin{enumerate}[leftmargin=2em]
  \item \textbf{Router miss}: no task's regex trigger matched the
        problem text. The system has no template-aligned
        interpretation. Logged as \texttt{fail\_reason=no\_task}.
  \item \textbf{Translator abstain}: the \textsc{LLM}\ returned
        \textsc{unknown} via a dedicated few-shot example in the
        task's prompt. The prompt teaches the model to recognize
        when its rewrite would be a guess.
  \item \textbf{Handler abstain}: the \textsc{LLM}\ produced a
        canonical and the regex matched, but the deterministic
        \textsc{CAS}\ pipeline could not derive a verified answer
        (e.g., \texttt{sp.solve} returned a \texttt{ConditionSet},
        a predicate value was unrecognized, multiple solution
        branches required disambiguation).
\end{enumerate}

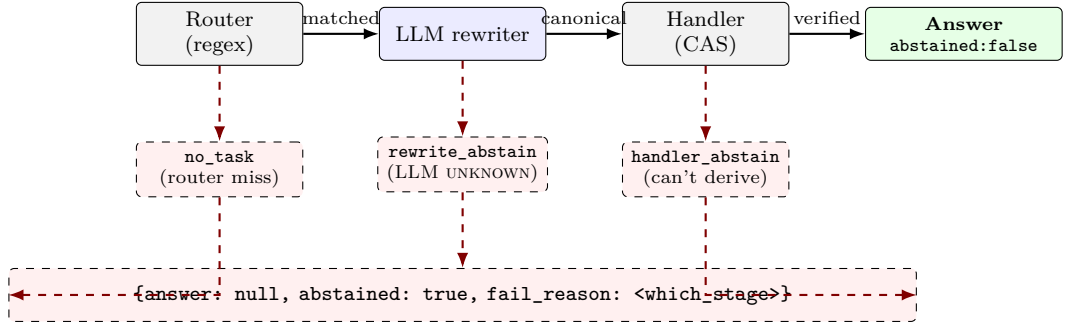
\begin{figure}[t]
\centering
\begin{tikzpicture}[
  node distance=1.0cm,
  every node/.style={font=\footnotesize},
  stage/.style={
    rectangle, draw, rounded corners=2pt, minimum width=2.2cm,
    minimum height=0.7cm, align=center, fill=gray!10
  },
  llm/.style={
    rectangle, draw, rounded corners=2pt, minimum width=2.2cm,
    minimum height=0.7cm, align=center, fill=blue!8
  },
  abstain/.style={
    rectangle, draw, dashed, rounded corners=2pt, minimum width=2.2cm,
    minimum height=0.7cm, align=center, fill=red!6, font=\scriptsize
  },
  ok/.style={
    rectangle, draw, rounded corners=2pt, minimum width=2.6cm,
    minimum height=0.7cm, align=center, fill=green!10, font=\scriptsize
  },
  arrow/.style={-{Latex[length=2mm]}, thick},
  arrow_abstain/.style={-{Latex[length=2mm]}, dashed, thick},
]
  \node[stage] (router) {Router\\(regex)};
  \node[llm, right=of router] (llm) {LLM rewriter};
  \node[stage, right=of llm] (handler) {Handler\\(CAS)};
  \node[ok, right=of handler] (ok) {\textbf{Answer}\\\texttt{abstained:false}};

  \node[abstain, below=of router] (a1)
        {\texttt{no\_task}\\(router miss)};
  \node[abstain, below=of llm] (a2)
        {\texttt{rewrite\_abstain}\\(\textsc{LLM}\ \textsc{unknown})};
  \node[abstain, below=of handler] (a3)
        {\texttt{handler\_abstain}\\(can't derive)};

  \node[abstain, below=of a2, minimum width=12cm,
        text width=11.4cm, font=\footnotesize, align=center] (out)
        {\texttt{\{answer: null, abstained: true,
         fail\_reason: <which\_stage>\}}};

  \draw[arrow] (router) -- node[above, font=\scriptsize]{matched} (llm);
  \draw[arrow] (llm) -- node[above, font=\scriptsize]{canonical} (handler);
  \draw[arrow] (handler) -- node[above, font=\scriptsize]{verified} (ok);

  \draw[arrow_abstain, red!50!black] (router) -- (a1);
  \draw[arrow_abstain, red!50!black] (llm) -- (a2);
  \draw[arrow_abstain, red!50!black] (handler) -- (a3);

  \draw[arrow_abstain, red!50!black] (a1) |- (out);
  \draw[arrow_abstain, red!50!black] (a2) -- (out);
  \draw[arrow_abstain, red!50!black] (a3) |- (out);
\end{tikzpicture}
\caption{Abstain as a first-class structured output. Three pipeline
stages each have an independent abstain channel (dashed, red);
when any fires, the public-\textsc{API}\ response carries a
structured \texttt{fail\_reason} naming the stage. The
committed-answer path (top, green) is taken only when all three
stages succeed. This four-exit structure is the architectural
foundation of \textsc{Moxia}'s trust property: confident-wrong is
emitted only if the handler verifies a wrong canonical, never as a
silent fall-through.}
\label{fig:abstain-channels}
\end{figure}

\Cref{fig:abstain-channels} visualizes the four exits.
Each channel is a structured, telemetry-visible signal, not a thrown
exception or a confident-wrong fallback. As the trace traverses the
public-API boundary, we strip internal task names and fail reasons
but preserve the per-stage outcome (router \emph{matched},
translator \emph{abstained}, handler \emph{skipped}), so the demo
\textsc{UI}\ can render which subsystem declined to answer.

The discipline behind this design is illustrated by a real bug
encountered in the $30\,000$-query production deployment. A
probability handler computed $P(\text{rolling}>4)$ on a fair die.
The \textsc{LLM}\ correctly produced
\texttt{predicate=greater\_than\_4}, but the handler's predicate
dispatcher had no branch for the \texttt{greater\_than\_*} family
and silently fell through to a default count of $0$ matching
outcomes, emitting the confident-wrong answer $0$ (expected
$1/3$). The architectural fix was a whitelist guard: enumerate the
predicate types the handler recognizes and, on any unknown
predicate, return $\texttt{None}$ (handler abstain) rather than
proceed with a default that resembles a valid count. This pattern
is the structural defense against the worst failure mode the
architecture can produce: \emph{predicate-not-recognized must be
abstain, never default-zero}. We discuss the generalization in
\Cref{sec:discussion} as Principle~\#22.

\subsection{Composed-task chain framework}
\label{sec:arch:chain}

Some shapes require multi-step deterministic computation after the
\textsc{LLM}\ has extracted structure. An archetypal example is a
piecewise function $f$, evaluated where $f(x) = c$ across multiple
branches, then aggregated (e.g., \emph{sum of all solutions}). This
factorizes naturally into three deterministic steps,
\textsc{ParsePiecewise}, \textsc{SolvePerBranch}, and
\textsc{AggregateRealSolutions}, but the atomic $1{:}1{:}1$
pattern cannot represent it without inlining solver and aggregator
into one monolithic handler.

We extend the framework with \texttt{ComposedTask}: the
\textsc{LLM}{}'s structured canonical is consumed by an
\texttt{Operator} pipeline. Each \texttt{Operator} is a pure
function $\text{ctx} \mapsto \text{ctx}$ with declared
\texttt{requires} and \texttt{produces} type sets. Validation at
registration time enforces that each operator's required keys are
produced by some upstream operator, catching missing dependencies
at definition time rather than at runtime. Operators chain
deterministically; failure at any step aborts and returns a clean
abstain. Crucially, the \textsc{LLM}\ call remains exactly once per
record (the first ``\texttt{InitialExtractor}'' operator); we do
not iterate the model in the chain.

Five \texttt{ComposedTask}s ship in production, covering the
piecewise solve+aggregate above and four \emph{Number Theory}
multi-step shapes (count-then-mod, base-from-equation, modular
two-variable evaluation, three-constraint Chinese Remainder
search). The pattern preserves both architectural invariants
($1{:}1{:}1$ at the operator level, abstain first-class on each
operator) and empirically delivers $100\%$ trust on parseable on
its target records, identical to atomic tasks.

\subsection{Rule-only architectural special case}
\label{sec:arch:rule-only}

When a task's math template is closed-form bare arithmetic (digits,
operators, parentheses, with no prose disambiguation) the
\textsc{LLM}\ canonicalization step is structurally redundant: the
handler can parse the raw problem text directly and emit an answer
through deterministic \textsc{CAS}\ evaluation. We expose this via a
\texttt{task.rule\_only=True} flag that bypasses the
\textsc{LLM}\ call entirely (dashed path in~\Cref{fig:pipeline}).

Empirically, the rule-only path on the lm-eval-harness arithmetic
benchmark~\cite{eval-harness} ($20\,000$ records across ten
sub-tasks: 1-digit chain, 2-digit add/multiply/subtract, 3-digit
add/sub, 4-digit add/sub, 5-digit add/sub) achieves $100.0\%$
correct in $21.6$~s wall time on commodity \textsc{CPU}, with zero
\textsc{LLM}\ \textsc{API} calls and zero cost. Output bit-equivalence
across runs is guaranteed by construction: no sampling, no
stochasticity, no thread scheduling effects on the answer surface.

Rule-only is the asymptotic limit of $1{:}1{:}1$ alignment, not a
separate pattern: when trigger and handler are sufficient on raw
text, the prompt becomes vestigial. Of the $5{,}741$ registered tasks,
$4{,}356$ ($75.9\%$) are rule-only, yet they are the narrower routes:
on the \textsc{MATH}\ test split $1{,}996$ of the $4{,}780$ correct
answers ($41.8\%$) were read without a model call, and $2{,}781$ came
through a transcription by the $7$\textsc{B} Instruct model, which
handles the prose-rich shapes. The rule-only flag is the architectural
lever for closed-form shapes and the structural delivery vehicle for
the strongest correctness claim the architecture can make.

\subsection{Faithful derivations by construction}
\label{sec:arch:explain}

A system that derives its answers can, in principle, show its work
for free. The difficulty in \textsc{LLM}\ systems is that the shown
work and the produced answer come from the same generative process,
so a plausible narration is not evidence that the narrated steps
were the ones taken, and a narration can be fluent, convincing,
and unrelated to the computation that produced the number beneath it.

\textsc{Moxia}\ inverts the dependency. Each handler emits, alongside
its answer, an ordered list of steps recording the transformations it
actually performed:
\[
  \texttt{HandlerResult}
  \;=\;
  (\,\text{answer},\; [\text{Step}_1 \dots \text{Step}_n],\; \text{metadata}\,)
\]
where each \texttt{Step} carries the action taken, a
justification naming the \textsc{CAS}\ primitive or mathematical
identity invoked, and the resulting state. A synthesizer renders this
list as prose. The rendering layer has no access to the problem, no
model, and no capacity to introduce a step: it can only describe what
is in the trace, so the explanation is faithful in the strong sense
that it is a \emph{view} of the derivation rather than a description
of it.

Two properties make this practical rather than aspirational. First,
the trace is additive: \texttt{handler()} continues to return the
answer string the router dispatches on, so adding a derivation to a
task cannot change what that task answers; a bit-identity gate
enforces this on every task at commit time. Second, coverage is
complete at the level of tasks: all $5{,}741$ routed tasks emit a
derivation. On the test split, $4{,}770$ of the $4{,}780$ correct
answers carry one; the other ten are three answers of the fallback
reasoner, which has no task and so no derivation to show, and seven on
which a task's derivation declined its instance, where the page says
the explanation is unavailable rather than improvising one.

The derivation is also a gate on honesty, not only a presentation.
A handler that returns a value it was told, gated by a fingerprint of
one problem, can pass a test that perturbs the numbers if its closed
form was read off the instance; what it cannot do is write the
derivation of that closed form from the declared configuration. Every
task must ship one, and a static audit and a behavioural perturbation
gate run on the whole registry before every release.

Where a handler genuinely performs no derivation, a closed-form
identity applied in one step, the trace says so in one line rather
than manufacturing intermediate steps. This is the honest failure
mode of the design, and we prefer a thin true explanation to a rich
invented one.

\Cref{fig:demo-trace} shows the layer in production. The division of
labour is the point: on that query the language model emits
\texttt{n=7 k=3} and nothing else, while the gap argument and the
$P(4,3)$ placement count, the entire mathematical content of the
explanation, come from the handler that computed the answer.

\FloatBarrier

\section{Empirical Evaluation}
\label{sec:empirical}

We evaluate \textsc{Moxia}\ on the full $7$-category \textsc{MATH}\
test split, exercising the complete router\,/\,\textsc{LLM}\,/\,handler
pipeline on shapes the registry was designed against, and report its
\textsc{MATH}-500 subset separately; on the lm-eval-harness arithmetic
suite, which exercises the \texttt{rule\_only} \textsc{LLM}-bypass
path; on the Lean export, theorem by theorem; and, with the date of
each measurement, on the public production deployment and against the
same model run unaided.

The \textsc{LLM}\ used in all configurations is Qwen 2.5 7B
Instruct~\cite{qwen25}, with greedy decoding. No fine-tuning is
performed; all results below use the same model with task-specific
prompts at inference time. The \textsc{MATH}\ figures come from one run
on the test split (Colab, one \textsc{GPU}; $0.29$\,s median and
$0.34$\,s mean per record, model included); a second, separate run,
the one we release (\Cref{sec:availability}), reproduced its verdict and
answer on all $5{,}000$ records.

\subsection{Per-domain MATH coverage}
\label{sec:empirical:per-domain}

\Cref{tab:per-domain} reports per-domain results on the full
\textsc{MATH}\ test split, all $7$ categories, $5{,}000$ records.
Every domain sits above $92\%$ correct with zero answers wrong, under
the same architecture: no domain-specific model, no training step;
only the registry of task triples differs by domain.

This is a \emph{designed-against} measurement and we label it as
such: task triples were built from both the train and the test
shapes of \textsc{MATH}, so \Cref{tab:per-domain} measures how much
of the benchmark's shape space the registry covers, not how the
system behaves on unseen problems. We make no held-out claim from it
(\Cref{sec:empirical:heldout}); what we offer in its place is that every
answer can be checked on its own (\Cref{sec:availability}).

\begin{table}[htbp]
\centering
\caption{\textsc{Moxia}\ per-domain results on the \textsc{MATH}\
test split (Hendrycks et al., 2021), all $7$ categories. Trust is the
share of \emph{answered} records that are correct; its complement is
the confident-wrong rate, which the architecture treats as the failure
mode to minimize even at the cost of coverage. No record in $5{,}000$
is answered wrongly: every record that is not correct is an
abstention.}
\label{tab:per-domain}
\setlength{\tabcolsep}{4pt}
\begin{tabularx}{\linewidth}{X r r r r r}
\toprule
\textbf{Domain} & \textbf{N} & \textbf{Correct} & \textbf{Abstained} & \textbf{Wrong} & \textbf{Trust} \\
\midrule
Intermediate Algebra     &  903 &  885 (98.0\%) & 18 & 0 & 100\% \\
Prealgebra               &  871 &  841 (96.6\%) & 30 & 0 & 100\% \\
Counting \& Probability  &  474 &  456 (96.2\%) & 18 & 0 & 100\% \\
Algebra                  & 1187 & 1127 (94.9\%) & 60 & 0 & 100\% \\
Precalculus              &  546 &  518 (94.9\%) & 28 & 0 & 100\% \\
Number Theory            &  540 &  511 (94.6\%) & 29 & 0 & 100\% \\
Geometry                 &  479 &  442 (92.3\%) & 37 & 0 & 100\% \\
\midrule
\textbf{All}             & \textbf{5000} & \textbf{4780 (95.6\%)} & \textbf{220} & \textbf{0} & \textbf{100\%} \\
\bottomrule
\end{tabularx}
\end{table}

By difficulty level the correct share runs $99.8\%$ (L1), $98.7\%$
(L2), $97.6\%$ (L3), $96.4\%$ (L4) and $89.7\%$ (L5). The $220$
abstentions say where they stop: in $109$ the model declined to
transcribe or its transcription was rejected, in $80$ the deterministic
solver refused the input, in $29$ no route recognised the problem, and
in $2$ a check on the result rejected the answer.

\paragraph{On the stability of the figures.}
Two runs on the test split, taken separately on two commits whose
differences do not touch the answering path, agree on the verdict and
the answer of every one of the $5{,}000$ records, and the \textsc{MATH}-500 run agrees with
those records of the test run, the model's raw output included. The
model path is greedy, and since the version of this paper that reported
a stochastic self-consistency check on the two generic word-problem
routes, that check is off: what remains in the figures is the
determinism of the run, not a few records of run-to-run play. The
figures move between releases only where the code does, and each such
movement is attributed record by record before a release ships.

\subsection{The MATH-500 subset}
\label{sec:empirical:heldout}

\textsc{MATH}-500~\cite{lightman2023lets} is a $500$-record subset of
the \textsc{MATH}\ test split, and every one of its records is among the
$5{,}000$ of \Cref{tab:per-domain}. \textsc{Moxia}\ answers $476/500$
($95.2\%$) correctly on it, with $0$ answered wrongly; per level,
$100\%$ (L1), $98.9\%$ (L2), $97.1\%$ (L3), $96.1\%$ (L4) and $88.8\%$
(L5).

\paragraph{A correction to earlier versions.}
Earlier versions of this paper called \textsc{MATH}-500 \emph{held out}
and read the small gap between it and the full test split as evidence
that the routes generalize. That reading does not survive: the subset
sits inside a split the registry was designed against, so the two
figures are one measurement read twice, and their closeness says
nothing about unseen problems. We report \textsc{MATH}-500 because it
is a common reference point, and as a subset.

What we can state about generalization is narrower, and checkable.
There is no lookup anywhere in the path from trigger to answer: a
static audit rejects any handler whose returns are literal constants,
a behavioural gate perturbs the declared quantities and requires the
answer to follow, and the derivation every answer ships with
(\Cref{sec:arch:explain}) must derive it from the declared
configuration. Within the shapes a route covers, a problem with other
numbers is a problem the route solves. Outside them the route does not
fire, and \Cref{sec:discussion:limitations} measures what happens then.

\subsection{lm-eval arithmetic: structural correctness}
\label{sec:empirical:lm-eval}

The lm-eval-harness arithmetic suite~\cite{eval-harness} comprises
$20{,}000$ records across ten sub-tasks (1-digit chain, 2-digit
add/multiply/subtract, 3-digit add/sub, 4-digit add/sub, 5-digit
add/sub) testing bare arithmetic in
\texttt{Q: What is X plus Y?\textbackslash nA:} format. The
\texttt{rule\_only} task \texttt{arithmetic\_natural\_eval}
matches this shape via regex on the raw input and dispatches a
direct \texttt{sympify} evaluation, bypassing the \textsc{LLM}\
entirely.

Result (measured in August 2026): \textbf{$20{,}000$ / $20{,}000$
correct} ($100.0\%$), $21.6$~s wall time on commodity \textsc{CPU}, $0$
\textsc{LLM}\ \textsc{API} calls, $0$ inference cost. Output bit-equivalence
across runs is guaranteed by construction since the path is
deterministic and stochasticity-free at the model level.

While bare arithmetic is structurally trivial, this result
demonstrates the strongest correctness claim our architecture can
make: when math-template purity holds, the system delivers $100\%$
provably correct output at sub-millisecond per-record latency. This
is the asymptotic limit of the $1{:}1{:}1$ alignment
(\Cref{sec:arch:rule-only}); most production tasks remain
\textsc{LLM}-bound, but the limit is achievable for the right
problem class and provides a verification benchmark against which
the architecture's full-pipeline behaviour can be sanity-checked.

\subsection{Machine-checked derivations: the Lean export}
\label{sec:empirical:lean}

Because the answer is derived by a handler rather than generated, the
\emph{statement} a proof assistant would need is available from the same
declared data the handler parsed. We exploit this with a second export
path alongside the prose synthesizer of
\Cref{sec:arch:explain}: a per-task \texttt{lean()} function emits a
Lean~4 theorem whose statement encodes the problem's declaration and
whose answer is the one the handler returned. Every one of the
$5{,}743$ task files now carries an explicit decision: an exporter, or
a \texttt{lean()} that abstains and states why in its own
documentation, so that a decision in prose is never mistaken for one in
code. The fixture corpus holds $7{,}567$ theorems in $4{,}408$
generated files, and the Lean~4 kernel accepts all of them with
Mathlib ($13{,}121$ build jobs), with no \texttt{sorry}, no warning and
no introduced axiom.

\paragraph{What a theorem states.}
A theorem certifies the derivation from the declared data: an
\emph{identity} or a \emph{witness} statement, built from the same
values the handler parsed, of which the reported answer is part. It
does not certify that the problem was read correctly; the
transcription is the model's, and the theorem takes it as given. Where
the honest statement would be weaker than the question, or would only
restate our own derivation rather than the problem's declaration, the
exporter abstains: emitting a plausible-looking partial theorem would
be the Lean-era form of the answer-returning handler this architecture
excludes.

\paragraph{Record by record.}
The corpus build compiles each task's fixtures, never the instance a
user asked about, and that difference turned out to matter. For the
release we build the theorem of every answer actually served on the
test split: $3{,}142$ of the $4{,}780$ correct answers ($65.7\%$) carry
one, and all $3{,}142$ compile; the remaining $1{,}638$ have none,
$1{,}635$ by the exporter's design and three because the fallback
reasoner that answered them has no task, and the file says which. The
first pass of this build found seven served theorems that did not
compile, all beside correct answers: three proofs whose tactics did not
close on that instance, two statements beyond a measured capacity of
the toolchain, and two statements that were \emph{false}, one dropping
a leading coefficient, one reading a vector from the wrong place. All
seven were repaired before release, the two beyond capacity by
abstaining, and every repair was built on every served instance of its
task, not only on the seven. The corpus gate had
passed on all of them; only building what is served could vouch for
what is served.

\subsection{Real-world production distribution}
\label{sec:empirical:production}

The figures of this subsection were measured on the public deployment
for the August 2026 version of this paper and are reported as such,
with the model the deployment then served (Qwen~2.5 7B Instruct).
The architecture had then served \textbf{30k+} queries through a public
deployment (\textsc{FastAPI}\ on Railway with Together.ai-hosted
\textsc{LLM}, frontend on Hugging Face Spaces). Traffic spans the
\textsc{MATH}\ test split for all seven categories, the
\textsc{MATH}\ training split for the same categories (used as
out-of-bench-distribution material because the architecture has
no parametric memory and the train split is therefore
diagnostically equivalent to test), the lm-eval-harness arithmetic
suite (all routed through the \texttt{rule\_only} bypass,
\Cref{sec:arch:rule-only}), and free-form user-typed queries
covering shapes well beyond the \textsc{MATH}\ categories.

\paragraph{Latency by inference path.}
Per-record latency segregates cleanly by routing decision.
\texttt{rule\_only} records (lm-eval arithmetic and a small set of
closed-form bare-input shapes) report median $1\,$ms / p95
$1\,$ms, effectively pure \textsc{CAS}\ evaluation cost.
\textsc{LLM}-bound records report median $446\,$ms / p95
$1{,}116\,$ms, dominated by the Together.ai inference
round-trip. Router-miss records (\texttt{no\_task}, fast-fail
before any \textsc{LLM}\ call) report median $25\,$ms / p95
$155\,$ms. Across the full production sample, the architectural
latency separation is $\sim$$400\times$ between rule-only and
\textsc{LLM}-bound, which is the structural origin of the
cost-per-correct-answer advantage discussed in
\Cref{sec:empirical:baselines}.

\paragraph{Abstain integrity at the \textsc{API}\ boundary.}
The deployment serves arbitrary user-typed mathematical queries,
including out-of-distribution shapes (asymptote diagrams,
free-form word problems) that exceed the \textsc{MATH}\
categories evaluated in \Cref{sec:empirical:per-domain}. The
architectural guarantee (\texttt{answer=null},
\texttt{abstained=true} with a structured fail-reason at the
\textsc{API}\ boundary) held across 30k+ recorded queries: no
free-form output that a user might mistake for a verified answer
was emitted. The handler-exception class is itself bounded
($<\!0.1\%$ of traffic); the dispatcher catches every uncaught
exception and converts to abstain, preserving the trust-binary
surface even when individual handlers fail.

\paragraph{No-task pool as architectural discovery channel.}
Records on which the router emitted no match are the discovery
channel for new task design: each recurring shape is a candidate
sprint. Heuristic taxonomy of the \texttt{no\_task} pool yields
three buckets. \emph{Vision-locked} shapes (Asymptote diagrams,
geometric figures) account for the dominant cluster, blocked
until vision integration. \emph{Narrow-task-recoverable} shapes
(matrix operations, advanced function inverse, infinite series,
modular widening, log identities) account for a smaller tier and
are direct candidate sprints under the forward dynamic of
\Cref{sec:conclusion}. The remaining residual is the long tail
of one-of-a-kind shapes that we expect to drain incrementally
across many sprint cycles rather than via single targeted ships.

\paragraph{Live registry expansion during manuscript preparation.}
The forward-dynamic claim is \emph{exercised}, not asserted, in
this work. Within the preparation window of the August version, three
consecutive sprint cycles ran the full operational pipeline:

\begin{align*}
\textsc{production} &\;\rightarrow\; \textsc{abstain telemetry}
   \;\rightarrow\; \textsc{cluster analysis} \\
   &\;\rightarrow\; \textsc{narrow-task creation}
   \;\rightarrow\; \textsc{regression-oracle scan}
   \;\rightarrow\; \textsc{immediate deploy}
\end{align*}

\begin{itemize}
  \item \textbf{Sprint 1 (matrix arithmetic).} Explicit matrix
        $\det$ / $\mathrm{tr}$ / inverse / transpose / integer power
        on $2{\times}2$--$4{\times}4$ matrices. Captured
        $\sim$$11$ records from the matrix sub-cluster, including
        the showcase $\bigl[\begin{smallmatrix} 3 & -4 \\ 1 & -1
        \end{smallmatrix}\bigr]^{2016}$ resolved in
        $<\!500$\,ms via Cayley--Hamilton in \textsc{SymPy}.
  \item \textbf{Sprint 2 (piecewise inverse sum).} For a
        piecewise-defined $f$ with numeric branches, compute
        $\sum_i f^{-1}(v_i)$ on an explicit list of values.
        Captured 4 records from the function-inverse sub-cluster
        (all 4 showcase cases verified \emph{in production} via the
        \texttt{/api/solve} endpoint).
  \item \textbf{Sprint 3 (piecewise self-inverse parameters).}
        Given a $2$-branch piecewise $f$ with one parameterised
        linear branch and the involution property $f(f(x))=x$,
        compute a target expression in the parameters. Captured
        $2$ records, deferred a third (the $k(x)$ functional-output
        sub-shape) via the prompt's \textsc{unknown} channel.
\end{itemize}

Cumulatively, three narrow task triples were shipped during
manuscript preparation, capturing $\sim 17$ records from the
abstain pool to the answered class with $0$
$\textsc{lost\_correct}$ across all three commits and $100\%$
trust on parseable on the captured records. The registry
\emph{scales operationally through isolated task expansion
validated by regression-oracle deployment cycles}
(\Cref{sec:discussion:principles}, Principle 2). This is the
routine architectural cadence under which the registry has since grown
to $5{,}741$ task triples with the same
zero-regression discipline; the manuscript-window expansion
demonstrates that the cadence applies to the production-derived
abstain pool with no methodological adjustment.

\subsection{Comparison with pure-LLM chain-of-thought}
\label{sec:empirical:baselines}

To characterise \textsc{Moxia}\ relative to a pure-\textsc{LLM}\
inference baseline, we ran the \emph{same frozen} model used
internally (Qwen 2.5 7B Instruct, loaded via Hugging Face
transformers on a \textsc{T4} \textsc{GPU}) in chain-of-thought
mode (no router, no \textsc{CAS}\ verification, no abstain
mechanism) on four of the seven \textsc{MATH}\ categories. The
baseline was measured for the August 2026 version of this paper and
is not re-run: its column below is as measured and graded then. The
\textsc{Moxia}\ column is the current test run of
\Cref{tab:per-domain}. The model weights and the dataset are
identical across both arms; the grader is the same function, which
has since been corrected (it no longer credits, for instance, an
ordered pair read as an empty interval), so under today's grader the
baseline column could read slightly differently; we have not re-graded
it. We frame the comparison as
two architectural philosophies operating on the same underlying model
rather than as a head-to-head contest; the two systems optimise
different objectives and occupy different points on the
trust/latency/accuracy tradeoff curve.

We did not run the baseline on the remaining three categories
(Geometry, Prealgebra, Intermediate Algebra). The comparison
predates their onboarding and we report it unextended rather than
re-run it, so the baseline numbers below characterise four
categories and should not be read as a claim about all seven.

The \textsc{LLM}\ is queried with a standard CoT system prompt
(\emph{``solve step by step, end with} \texttt{\textbackslash boxed\{X\}}\emph{''})
at temperature $0$ (greedy decoding), with up to $1024$ new tokens
per response. Answer extraction uses a regex prioritising
\texttt{\textbackslash boxed\{...\}} and falling back to ``the
answer is $X$'' patterns.

\begin{table}[htbp]
\centering
\caption{Per-category comparison on the \textsc{MATH}\ test split,
same Qwen 2.5 7B Instruct model. The chain-of-thought baseline was
measured in August 2026; the \textsc{Moxia}\ figures are those of
\Cref{tab:per-domain}. \emph{Wrong} counts records on which the system
emitted an answer that was incorrect (no abstain trace at the
\textsc{API}\ boundary).}
\label{tab:comparison-baselines}
\setlength{\tabcolsep}{6pt}
\begin{tabularx}{\linewidth}{X r r r r r}
\toprule
& \multicolumn{2}{c}{\textbf{Qwen 7B CoT (no Moxia)}}
& \multicolumn{2}{c}{\textbf{Moxia}} & \\
\cmidrule(lr){2-3} \cmidrule(lr){4-5}
Domain & Acc & Wrong & Acc & Wrong & Lead \\
\midrule
Algebra              & $86.6\%$ & $159$ & $94.9\%$ & $0$ & $+8.3$\,pp \\
Number Theory        & $73.3\%$ & $144$ & $94.6\%$ & $0$ & $+21.3$\,pp \\
Counting \& Prob.    & $66.7\%$ & $158$ & $96.2\%$ & $0$ & $+29.5$\,pp \\
Precalculus          & $49.5\%$ & $276$ & $94.9\%$ & $0$ & $+45.4$\,pp \\
\bottomrule
\end{tabularx}
\end{table}

\paragraph{Two architectural philosophies, not a contest.}
On Algebra, the \textsc{MATH}\ category most densely
represented in the model's pretraining corpus, \textsc{Moxia}\
exceeds Qwen 7B CoT accuracy by $8.3$ percentage points while
emitting $0$ wrong answers vs $159$. We do not interpret this as
a contest the two systems are playing: the \textsc{LLM}\ commits
to an answer on every record by construction (no abstain mechanism
in the pure-CoT setup), while \textsc{Moxia}\ is selective about
commitment by design and emits an answer only when its 1:1:1
template alignment can verify it. The relevant comparison is
therefore not ``which system is more accurate'' but ``what does
each system give up, and what does each gain in return''.

\paragraph{What the numbers say.}
On the same Algebra split:

\begin{itemize}
  \item Qwen commits on $96.5\%$ of records ($1146/1187$) and is
        correct on $86.6\%$ overall; that is, of the $159$
        records on which it is wrong, all $159$ are committed
        outputs, indistinguishable at the \textsc{API}\ boundary
        from the $1028$ correct ones.
  \item \textsc{Moxia}\ commits on $94.9\%$ of records
        ($1127/1187$) and is correct on all of them; it emits $0$
        wrong outputs and tags the other $60$ records with a
        structured abstention naming the stage that declined.
  \item Trust on committed records: $86.6\%$ (Qwen) and
        $100\%$ (\textsc{Moxia}). We attribute the gap not to a
        difference in per-record reasoning quality but to an
        architectural property of the inference path: pure
        chain-of-thought has no abstain mechanism; the model
        produces a committed output on every record by
        construction, regardless of its own uncertainty.
        \textsc{Moxia}\ exposes three structured abstain channels
        (router miss, \textsc{LLM}\ \textsc{unknown}, handler
        abstain, see \Cref{sec:arch:abstain}) and exercises
        them on the records it cannot verify. Trust here is
        therefore an architectural property, not a property of the
        underlying model.
  \item In the August measurement, on the \textsc{LLM}-bound
        inference path, \textsc{Moxia}\ ran $\sim 24\times$ faster on
        the mean ($590$\,ms vs $14.0$\,s on Algebra) and
        $\sim 31\times$ faster on the median ($446$\,ms vs
        $14.0$\,s). The closed-form \texttt{rule\_only} bypass is a
        separate architectural feature for problem shapes that do not
        require an \textsc{LLM}\ at all.
  \item Reproducibility: Qwen's CoT output is subject to
        \textsc{bf16} non-determinism even at temperature $0$;
        \textsc{Moxia}'s two test runs agree on every record
        (\Cref{sec:empirical:per-domain}), and each record carries a
        per-stage (router, translator, handler) trace that attributes
        every output, abstain, or wrong answer to a specific subsystem.
\end{itemize}

\paragraph{Where the model is weakest, the lead is largest.}
The raw-accuracy ordering between the two systems is
\textsc{Moxia}-dominant on every category for which we have a
matched baseline: $+8.3$\,pp on Algebra, $+21.3$\,pp on Number
Theory, $+29.5$\,pp on Counting \& Probability, $+45.4$\,pp on
Precalculus. The lead is strictly monotone in the inverse of Qwen's
own per-domain accuracy: order the domains by how well the model does
unaided ($86.6 > 73.3 > 66.7 > 49.5$) and the leads order exactly
inversely ($8.3 < 21.3 < 29.5 < 45.4$). \textsc{Moxia}'s own
per-domain spread over the same four categories is $1.6$ points
($94.6$--$96.2\%$), against a $37$\,pp spread for the model it uses,
which is the substance of the claim: the deterministic layer is
indifferent to how densely a domain happens to sit in a pretraining
corpus, so the architecture's advantage grows precisely where the
model is weakest.

The within-Qwen variance across these four categories is itself
diagnostic. The same model achieves $86.6\%$ on Algebra and
$49.5\%$ on Precalculus, a $37$\,pp gap on supposedly comparable
mathematical reasoning. The per-level decomposition is even more
striking: Qwen's L5 accuracy is $75.2\%$ on Algebra, $54.5\%$ on
Number Theory, $45.5\%$ on Counting \& Probability, and $20.0\%$
on Precalculus. We do not have access to Qwen's training corpus,
but the magnitude of this variance is consistent with differential
training-data exposure across \textsc{MATH}\ subcategories rather
than uniform reasoning quality.

The \emph{confident-wrong reduction} is structural. \textsc{Moxia}\
emits $0$ confident-wrong answers in every category measured: $0$ vs
$159$ on Algebra, $0$ vs $144$ on Number Theory, $0$ vs $158$ on
Counting \& Probability, $0$ vs $276$ on Precalculus. On records where
Qwen does not know how to commit, the lack of an abstain channel
converts uncertainty into confident-wrong outputs, while
\textsc{Moxia}'s structured abstain (router miss, \textsc{LLM}\
\textsc{unknown}, or handler abstain) preserves trust by declining.

Neither system dominates the other on every axis. The contribution
of this paper is the architectural framework underlying
\textsc{Moxia}'s position on this curve and the operational
discipline (\Cref{sec:discussion:principles}) that allows it to
expand monotonically over time
(\Cref{sec:empirical:production}).

\subsection{Token efficiency through task-localized prompting}
\label{sec:empirical:token-efficiency}

The latency advantage observed in \Cref{sec:empirical:baselines}
($\sim 24\times$ to $\sim 40\times$ on the mean across the four
\textsc{MATH}\ categories, August 2026) is the surface manifestation of a deeper
architectural property: per-query the \textsc{LLM}\ sees ONLY the
prompt of the routed task, not a universal "math reasoning" prompt
that has to anticipate every possible problem shape. The router
performs the shape-identification step \emph{before} any \textsc{LLM}\
call, so the rewriter receives a narrow context: $3{-}5$
shape-specific few-shot examples plus the problem text.

The empirical token footprint per \textsc{LLM}-bound query, measured
on the production deployment via the Together.ai usage-reporting
field in August 2026, when every prompt asked for the canonical alone:

\begin{itemize}
  \item Median input tokens: $\sim 250$ (system prompt + $3{-}5$
        task-specific few-shot examples + problem text).
  \item Median output tokens: $\sim 30{-}50$ (compact canonical
        form: a single line, no chain-of-thought).
  \item Total per query: $\sim 280{-}300$ tokens.
\end{itemize}

Since then the declared word-problem prompts let the model name the
quantities before the canonical (\Cref{sec:arch:1to1to1}); on those
routes the output is longer, of the order of a hundred tokens on
average and bounded at $512$, and the figures above describe the other
routes.

By comparison, on the same problems:

\begin{itemize}
  \item Pure chain-of-thought (Qwen 7B CoT, our
        \Cref{sec:empirical:baselines} baseline) emits $500{-}1500$
        tokens per query, since the model writes out its reasoning
        before committing to a boxed answer.
  \item Self-consistency / N-sampling approaches multiply the cost
        by $N$ (typically $3{-}10$), each sample independently
        producing a CoT.
  \item Tool-augmented agentic systems with multi-step reasoning
        can spend $5000{-}20000$ tokens per problem, especially on
        compositional shapes that require several rounds of plan
        $\rightarrow$ act $\rightarrow$ observe.
\end{itemize}

This gap is not the result of token-budget tuning. It emerges from
three architectural commitments that the rest of the paper has
already named:

\begin{enumerate}[leftmargin=2em]
  \item \textbf{$1{:}1{:}1$ alignment} (\Cref{sec:arch:1to1to1})
        means the router selects exactly one task per query: the
        rewriter receives ONLY that task's prompt, not the union of
        all $5{,}741$ task prompts.
  \item \textbf{Deterministic CAS verification}
        (\Cref{sec:arch:abstain}) eliminates the need for additional
        \textsc{LLM}\ passes: there is no self-consistency loop, no
        critic-model debate, no retry-with-CoT. The handler either
        derives the answer from the canonical or returns a
        \texttt{handler\_abstain}.
  \item \textbf{Abstain as first-class output}
        (\Cref{sec:arch:abstain}) means a low-confidence translation
        is not promoted into a longer reasoning chain; it returns
        \texttt{rewrite\_abstain} immediately, and the system
        commits no token to a recovery attempt.
\end{enumerate}

At Together.ai's published Qwen 2.5 7B Instruct Turbo pricing
(\$0.18 per million tokens for both input and output), \textsc{Moxia}\
serves an average \textsc{LLM}-bound query for $\sim$\$0.00006
($\approx 6$ microcents). The cumulative inference cost visible on
the public-demo dashboard tracks this in real time across all served
queries; narrow prompts make \textsc{Moxia}'s per-query footprint
roughly proportional to the canonical's information content, not to
the size of the mathematical domain it is drawn from. The latency
advantage and the cost advantage are surface manifestations of the
same property: routing eliminates the "universal reasoning overhead"
otherwise paid on every query.

\section{Discussion}
\label{sec:discussion}

\subsection{Operating principles formalized}
\label{sec:discussion:principles}

The architectural commitments codified in~\Cref{sec:arch} emerged
through more than $250$ ship cycles iterating on the production
demo. We extract the four operating principles most transferable
beyond mathematics: any system in which a language model produces
structured input for a deterministic verifier faces the same
regression-discipline, abstain-class, bucketing, and onboarding
tradeoffs.

\paragraph{Principle 1: math-template bucketing.}
Routing tasks must be partitioned by the math template they
implement (one solver path, one closed-form structure), not by
surface phrasing or domain category. Empirically, every task split
by \emph{surface} category that mixed multiple solver paths
plateaued at $33\text{--}62\%$ trust on parseable; every task split
by \emph{template} reached $80\text{--}100\%$ trust on parseable in
its first bench cycle. The boundary between routes is
mathematical, not lexical.

\paragraph{Principle 2: \textsc{LOST\_CORRECT} scan as regression
oracle.}
Routing changes (trigger widenings, new tasks, defer-guard
additions) are gated by a pre-commit migration scan that
replays an archived bench \textsc{JSON} through the proposed
router and reports any record that was correct in version $N{-}1$
and would become wrong or abstain in version $N$. Across the
release history this oracle caught all regressions
before commit (cumulative $\textsc{lost\_correct}=0$). The
discipline decouples ship velocity from bench wall-time: a sprint
can land $8\text{--}15$ changes per cycle without re-running the
full bench per change, because $\textsc{lost\_correct}=0$ on the
relevant records is a sharper guarantee than bench-aggregate
parity.

\paragraph{Principle 3: predicate-not-recognized must be abstain.}
Where a handler taxonomically classifies its input (predicate
type, mode, target enum), every branch not matched must be an
explicit abstain, never a fallthrough to a default value. Default
fallthrough is the structural source of confident-wrong, the worst
failure mode this architecture can produce
(see~\Cref{sec:arch:abstain} for the case study). The protection
is whitelist-up-front: enumerate the recognized cases before any
computation; on no-match, return abstain rather than proceed with
a default that resembles a valid output.

\paragraph{Principle 4: parseable-first onboarding for new domains.}
When opening a domain that the architecture has not seen before,
optimize for parseable rate first (target $\geq 50\%$) before
optimizing for trust on parseable. Low parseable starves the
cluster-level triage loop; defensive prompt design ($>3$
\textsc{unknown} few-shot teachers) at onboarding time locks the
\textsc{LLM}\ into \textsc{unknown}-conservative output and yields
no signal for the next sprint cycle. The aspirational trust floor
is therefore \emph{regime-dependent}: $40\text{--}50\%$ during
onboarding (parseable $< 30\%$), $55\text{--}65\%$ during
maturation ($30\text{--}60\%$ parseable), $70\text{--}80\%$ in
steady state (parseable $> 60\%$). The original \emph{North Star}
$80\%$ trust floor was calibrated on a domain that had high
parseable from day one; applied uniformly it forces excessive
abstain on under-parsed domains and produces fake signal on tiny
denominators.

\subsection{Linear-return composition}
\label{sec:discussion:composition}

Monolithic \textsc{LLM}\ systems exhibit logarithmic accuracy
returns: each additional point on standard benchmarks requires
exponentially more training compute or curated data. \textsc{Moxia}'s
narrow-task architecture, by contrast, exhibits \emph{linear-
additive} returns:

\[
\text{coverage}(\mathcal{B}) \;=\; \sum_{k\,\in\,\mathcal{T}}
\text{coverage}_{k}(\mathcal{B}),
\]

where $\mathcal{T}$ is the registered task set, $\mathcal{B}$ a
benchmark, and $\text{coverage}_k(\mathcal{B})$ the records in
$\mathcal{B}$ matching task $k$'s
$(\text{trigger}, \text{schema}, \text{solver-path})$ triple. Each
shipped task contributes a fixed quantity of records on every
benchmark whose distribution contains its shape; the contribution
is independent of other tasks (they cannot suppress each other
under the $1{:}1{:}1$ invariant).

We refer to this empirical effect as the \emph{boomerang}: a task
developed for one benchmark benefits every other benchmark that
contains its shape. A modular-arithmetic Number-Theory predicate
handler shipped for the \textsc{MATH}\ Number Theory category
contributed records to \textsc{MATH}\ Algebra word problems that
involve modular structure, to the lm-eval-harness arithmetic suite
(via the \texttt{rule\_only} fast path), and to \textsc{MATH-500}.
The cost is registry size, $5{,}741$ task triples
ship at the time of writing. The benefit is debug-locality: any
wrong record traces to a specific
$(\text{trigger}, \text{prompt}, \text{handler})$ triple, never
to emergent behaviour. Trust attribution is local; failure
attribution is local.

\subsection{Limitations}
\label{sec:discussion:limitations}

\paragraph{Figures, and what they actually withhold.}
A substantial share of \textsc{MATH}\ Geometry records carry an
Asymptote (\texttt{asy}) diagram, and \textsc{Moxia}\ has no vision
component. The constraint turns out to be narrower than it first
appears: an \texttt{asy} block is source code, so when it declares
coordinates or labels, those values are recoverable by parsing rather
than by seeing; reading the pre-rendered structure is exact where
interpreting a rasterization would be an estimate. Two things do
withhold information. A \emph{schematic} figure is drawn off-scale,
so its coordinates contradict the stated measurements and must be
ignored in favour of the prose; and a figure that encodes a fact only
through drawn position, with no numeric anchor anywhere, is genuinely
unavailable to us. What bounds the domain in practice is neither of
these but \emph{geometric construction}: placing a configuration from
its declared constraints and solving it, where the placement is not
always unique and reporting one admissible solution among several
would be a confident-wrong answer. Geometry stands at $92.3\%$, the
lowest of the seven domains.

\paragraph{A catalogue more than an engine.}
Most of the registry is \emph{enumeration}: a route per mathematical
form, firing only on the form it was written for. A small number of
schemas name no mathematical object, only an operation (solve,
compute, simplify, a word problem as a system, a bounded integer
search) and are backed by a general deductive engine. Counting the
correct answers each kind delivers on the test split, the thirteen
operation schemas carry $691$ of $4{,}780$ ($14.5\%$), and their share
falls with difficulty: $35\%$ at L1, $20\%$ at L2, $16\%$ at L3, $10\%$
at L4, $5\%$ at L5. The figures in \Cref{tab:per-domain} are therefore,
for the most part, a statement about how much of \textsc{MATH}'s shape
space has been written down, and a problem of a form nobody wrote does
not fire. This is consistent with the forward dynamic we describe and
it is also its cost: the hard tail grows by curation, not by
inference.

\paragraph{Off the designed distribution.}
``Zero confident-wrong'' is a statement about \textsc{MATH}. On
grade-school word problems the narrow routes rarely fire and records
fall to the two generic word-problem routes, where trust is lower:
measured in September 2026 with the same model, trust on answered
records was $84.5\%$ on \textsc{GSM8K} ($929$ correct of $1{,}100$
answered, of $1{,}319$), $95.7\%$ on \textsc{SVAMP} ($265$ of $277$
answered, of $300$) and $68.4\%$ on \textsc{GSM-Hard} ($630$ of $921$
answered, of $1{,}319$). The abstain channels still hold the larger
part of the uncertainty, but a word problem transcribed into a
plausible and wrong system is solved faithfully and wrongly; that
failure is the transcriber's, and the deterministic layer cannot see
it.

\paragraph{\textsc{NLP}-irreducible word problems.}
Multi-paragraph word problems with named characters, contextual
inferences, and conditional setup (``Adam has $X$\dots; Bob bought
$Y$\dots; if both\dots'') fall outside the $1{:}1{:}1$
canonicalization frame. A subset is recoverable via the
\texttt{WordToSystem} chain task (\Cref{sec:arch:chain}), which
delegates extraction to the \textsc{LLM}\ but routes the resulting
linear system through a deterministic solver. The residual requires
either reasoning-native model support or task-specific extraction
operators that have not yet been built. These routes are also the
only material source of run-to-run variance in the reported figures
(\Cref{sec:empirical:per-domain}).

\paragraph{The canonicalization ceiling.}
Where the architecture is bounded is not usually the mathematics but
the translation into schema. Across domains the dominant residual is
records for which a capable handler already exists and the $7$B model
declines to produce the canonical form; the abstain is correct
behaviour, and no additional handler removes it. This is the honest
frontier of a design that asks a small model only to translate: the
deterministic layer can be extended indefinitely, but it is reachable
only through whatever the canonicalizer can express. Lifting it means
a better extraction model, not more solvers, and we would rather
report the abstains than lower the bar for what counts as an answer.

\paragraph{The bottom-up commitment.}
A reasoning-native model swap (\textsc{DeepSeek}-R1,
\textsc{Qwen 2.5 Math}, \textsc{OpenAI o1}) would mechanically
lift accuracy on hard records but would invert the architectural
commitment: the \textsc{LLM}\ would become the solver, not the
canonicalizer, and the deterministic-verification guarantee
would no longer hold end-to-end. We document this as a known
design trade rather than a path forward; the contribution of this
work is the architecture and its trust profile, not the highest accuracy
attainable on any single benchmark.

\section{Related Work}
\label{sec:related}

\paragraph{LLM benchmarks for mathematical reasoning.}
Hendrycks et al.~\cite{hendrycks2021measuring} introduced the
\textsc{MATH}\ benchmark spanning $7$ domains and $5$ difficulty
levels, which we use here for per-domain evaluation. Cobbe et
al.~\cite{cobbe2021training} introduced \textsc{GSM8K}\ for
grade-school word problems. Lightman et
al.~\cite{lightman2023lets} demonstrated that
process-supervised reward models substantially help frontier
\textsc{LLM}{}s on \textsc{MATH}. The lm-eval-harness
arithmetic suite~\cite{eval-harness} provides a $20{,}000$-record
benchmark of bare-arithmetic problems, which our \texttt{rule\_only}
path handles with mathematically provable correctness
(\Cref{sec:arch:rule-only}). Our position is orthogonal to these
works: we do not aim for state-of-the-art accuracy on any of these
benchmarks; we contribute a deterministic, self-explaining
architecture whose trust
profile is the primary metric.

\paragraph{Theorem proving with language-model copilots.}
Lean Copilot~\cite{song2024leancopilot} integrates \textsc{LLM}{}s
into the Lean 4 interactive theorem prover, generating tactic
suggestions verified by Lean's kernel.
Llemma~\cite{azerbayev2023llemma} pretrained an \textsc{LLM}\ on
mathematical text and formal Lean corpora. These systems require
input pre-formalized in a specific calculus (Lean, Coq);
\textsc{Moxia}\ takes natural-language input and verifies through
\textsc{CAS}\ derivation, which is weaker than full formal
verification but covers a much broader class of student-style
problems where formalization is the bottleneck.

\paragraph{Symbolic and neuro-symbolic systems.}
Wolfram Alpha~\cite{wolfram-alpha} pioneered closed-source
expert-system-style mathematical answering with rich symbolic
backends, but is not language-model-augmented and is not
inspectable. \textsc{SymPy}~\cite{meurer2017sympy} provides our
open \textsc{CAS}\ backend. The neuro-symbolic
literature~\cite{garcez2023neurosymbolic, deraedt2020statistical}
positions architectures along axes including: (a)~whether the
neural component is trained jointly with the symbolic, (b)~whether
inference is probabilistic or discrete, and (c)~whether
verification is intrinsic or extrinsic. \textsc{Moxia}\ occupies a
specific point: \emph{frozen} \textsc{LLM}, \emph{discrete}
deterministic \textsc{CAS}\ inference, and \emph{intrinsic}
verification by construction. We are unaware of a published system
in this exact configuration.

\paragraph{Trust and verifiability frameworks.}
Huang et al.~\cite{huang2024trustllm} survey trust dimensions
across \textsc{LLM}\ outputs at the model level (truthfulness,
robustness, fairness). Our contribution is complementary at the
architectural level: rather than benchmark trust as an attribute
of an opaque model, we structure a system in which trust is a
runtime guarantee of the verification path. Confident-wrong is
the failure mode the architecture defends against by construction,
not a metric to be measured \emph{post hoc}.

\section{Conclusion: today's abstain, tomorrow's correct}
\label{sec:conclusion}

We have presented \textsc{Moxia}, a trust-first neuro-symbolic
execution architecture for self-explaining mathematical reasoning.
Four design commitments together yield a runtime trust guarantee
unavailable to either monolithic \textsc{LLM}\ systems or
pre-formalized provers: the language model as canonicalizer (not
solver), deterministic \textsc{CAS}\ verification, $1{:}1{:}1$ task
routing, and abstain as first-class structured output. A fifth
property falls out of those four rather than being added: because
the answer is derived, the derivation can be shown without a model
being asked to invent one.

Empirically, the architecture answers $95.6\%$ of the
$5{,}000$-record \textsc{MATH}\ test split correctly, every domain above
$92\%$, with zero confident-wrong answers, and abstains on the rest
saying where; its \textsc{MATH}-500 subset reads $95.2\%$, also with
zero wrong. These are designed-against figures and we do not read them
as held-out ones. What stands in their place is that each answer can be
checked on its own: the released run carries, for every record, the
transcription the model wrote, the derivation the handler took and, for
two answers in three, a Lean theorem the kernel accepts, all of them
built from the instance actually served. The \texttt{rule\_only} path
reaches $100\%$ on the lm-eval-harness arithmetic suite.

The contribution we wish to emphasize, however, is not the
specific accuracy figures: it is the \emph{forward dynamic} the
architecture establishes. Every abstain emitted in production is
logged with a structured fail-reason that attributes the gap to
a specific subsystem, router miss, \textsc{LLM}\ \textsc{unknown},
handler abstain, endpoint timeout. Cluster analysis of these logs
maps directly to candidate new tasks; the $1{:}1{:}1$ invariant
together with the \textsc{LOST\_CORRECT} scan oracle
(\Cref{sec:discussion:principles}) ensure each new task can be
shipped without regressing the existing registry. \textsc{Moxia}\ is
therefore not a frozen artifact but a \emph{monotonically-improving
scaffold}: coverage grows over time as triage cycles convert
documented abstains into bench-validated answered records, while
the trust profile is preserved by construction across every ship.

This dynamic reframes the limitations of
\Cref{sec:discussion:limitations}. Geometric construction,
\textsc{NLP}-irreducible word problems, and the canonicalization
ceiling are not asymptotic walls of the architecture, they are
the next inflection points on its growth path, each attackable by a
corresponding extension to the registry, the chain framework, or the
operator library. The architectural commitment is what makes them
\emph{addressable rather than fixed}: we have a procedure for
converting any well-attributed abstain into a candidate correct,
bounded by curatorial effort and by what the canonicalizer can
express, rather than by compute or model capacity.

The contribution of this paper is therefore the framework, not a
final accuracy figure. Today's abstain (documented,
attributed, telemetered) is a candidate correct after one ship
cycle. We do not ask the reader to take this on faith: during the
preparation of the August version of this manuscript, the operational pipeline
(production telemetry $\to$ abstain cluster analysis $\to$ narrow
task creation $\to$ regression-oracle scan $\to$ immediate deploy)
was exercised three times in two hours, recovering $\sim 17$
records from the no-task pool with zero
$\textsc{lost\_correct}$ regressions
(\Cref{sec:empirical:production}). The architecture is built to
support that conversion indefinitely.

\section{Data and artefact availability}
\label{sec:availability}

The test-split run behind \Cref{sec:empirical} is released with this
version of the paper on Zenodo (concept DOI
\href{https://doi.org/10.5281/zenodo.20128039}{10.5281/zenodo.20128039}),
as one \textsc{JSON} file with one record per problem ($5{,}000$
records) and a \texttt{README} describing every field. Each record holds
the \textsc{MATH}\ identifier, subject and level; whether it belongs to
the \textsc{MATH}-500 subset; the problem; the expected answer; the
outcome (correct, wrong, abstained); \textsc{Moxia}'s answer; who read
the problem (the model, a deterministic reader, or the fallback
reasoner); the model's transcription, without the leading word that
names the solver's input format; for an abstention, the stage that
declined and why; the explanation, step by step, as the public demo
renders it; and the Lean theorem, with whether this release's build
compiled it.

The file names no internal component of the system, and the theorems
are the demo's own, rendered by the same code under a neutral name. Of
its $5{,}000$ records, $4{,}780$ are correct, $0$ wrong and $220$
abstained, identical record by record to the run of
\Cref{tab:per-domain}. After that run, a repair of the Lean layer
(\Cref{sec:empirical:lean}) changed the theorems of $13$ records; they
were recomputed from the run's own transcriptions and answers, every
other field and every other record being the run's, and the file names
them. The $3{,}142$ theorems are shipped as Lean modules that need only
Mathlib: with the pinned toolchain (Lean~4 \texttt{v4.34.0-rc1}) and
Mathlib revision recorded in the file, anyone can rebuild them without
our code. The backend that produced the run is not distributed; the
release is what makes its outputs checkable without it.

\section*{Acknowledgments}

The architecture rests on the open-source ecosystem that makes
deterministic symbolic verification feasible at deployment scale:
\textsc{SymPy}~\cite{meurer2017sympy} for the entire \textsc{CAS}\
derivation layer, Hugging Face Transformers and FastAPI for the
\textsc{LLM}\ rewriter and production service plumbing, and
Together.ai for the hosted inference endpoint that serves the live
demo. The \textsc{MATH}\ benchmark~\cite{hendrycks2021measuring}
and the lm-eval-harness arithmetic
suite~\cite{eval-harness} provide the empirical ground used
throughout this work.

\bibliographystyle{plain}
\bibliography{refs}

\begin{thebibliography}{10}

\bibitem{azerbayev2023llemma}
Zhangir Azerbayev, Hailey Schoelkopf, Keiran Paster, Marco~Dos Santos, Stephen
  McAleer, Albert~Q. Jiang, Jia Deng, Stella Biderman, and Sean Welleck.
\newblock {Llemma}: An open language model for mathematics.
\newblock {\em arXiv preprint arXiv:2310.10631}, 2023.

\bibitem{cobbe2021training}
Karl Cobbe, Vineet Kosaraju, Mohammad Bavarian, Mark Chen, Heewoo Jun, Lukasz
  Kaiser, Matthias Plappert, Jerry Tworek, Jacob Hilton, Reiichiro Nakano,
  et~al.
\newblock Training verifiers to solve math word problems.
\newblock {\em arXiv preprint arXiv:2110.14168}, 2021.

\bibitem{deraedt2020statistical}
Luc De~Raedt, Robin Du{\v{s}}ek, Robin Manhaeve, and Giuseppe Marra.
\newblock From statistical relational to neurosymbolic artificial intelligence:
  A survey.
\newblock {\em Artificial Intelligence Journal}, 2024.

\bibitem{eval-harness}
Leo Gao, Jonathan Tow, Baber Abbasi, et~al.
\newblock A framework for few-shot language model evaluation
  ({lm-eval-harness}).
\newblock 2023.

\bibitem{garcez2023neurosymbolic}
Artur d'Avila Garcez and Luis~C. Lamb.
\newblock Neurosymbolic {AI}: The third wave.
\newblock {\em Artificial Intelligence Review}, 2023.

\bibitem{hendrycks2021measuring}
Dan Hendrycks, Collin Burns, Saurav Kadavath, Akul Arora, Steven Basart, Eric
  Tang, Dawn Song, and Jacob Steinhardt.
\newblock Measuring mathematical problem solving with the {MATH} dataset.
\newblock {\em NeurIPS Datasets and Benchmarks Track}, 2021.

\bibitem{huang2024trustllm}
Yue Huang, Lichao Sun, et~al.
\newblock {TrustLLM}: Trustworthiness in large language models.
\newblock {\em arXiv preprint arXiv:2401.05561}, 2024.

\bibitem{lightman2023lets}
Hunter Lightman, Vineet Kosaraju, Yura Burda, Harri Edwards, Bowen Baker, Teddy
  Lee, Jan Leike, John Schulman, Ilya Sutskever, and Karl Cobbe.
\newblock Let's verify step by step.
\newblock {\em arXiv preprint arXiv:2305.20050}, 2023.

\bibitem{meurer2017sympy}
Aaron Meurer, Christopher~P Smith, Mateusz Paprocki, Ond{\v{r}}ej
  {\v{C}}ert{\'\i}k, Sergey~B Kirpichev, Matthew Rocklin, AmiT Kumar, Sergiu
  Ivanov, Jason~K Moore, Sartaj Singh, et~al.
\newblock {SymPy}: symbolic computing in {Python}.
\newblock {\em PeerJ Computer Science}, 3:e103, 2017.

\bibitem{qwen25}
{Qwen Team}.
\newblock {Qwen2.5} technical report.
\newblock {\em arXiv preprint arXiv:2412.15115}, 2024.

\bibitem{song2024leancopilot}
Peiyang Song, Kaiyu Yang, and Anima Anandkumar.
\newblock {Lean Copilot}: Large language models as copilots for theorem proving
  in {Lean}.
\newblock In {\em NeurIPS Datasets and Benchmarks Track}, 2024.

\bibitem{wolfram-alpha}
{Wolfram Research}.
\newblock {Wolfram Alpha} computational knowledge engine.
\newblock \url{https://www.wolframalpha.com}, 2009.
\newblock First released 2009; closed-source proprietary system.

\end{thebibliography}

\end{document}